# Empowering Visually Impaired Individuals: A Novel Use of Apple Live Photos and Android Motion Photos

Seyedalireza Khoshsirat    Chandra Kambhamettu

*Video/Image Modeling and Synthesis (VIMS) Lab, University of Delaware*

**Abstract**

Numerous applications have been developed to assist visually impaired individuals that employ a machine learning unit to process visual input. However, a critical challenge with these applications is the sub-optimal quality of images captured by the users. Given the complexity of operating a camera for visually impaired individuals, we advocate for the use of Apple Live Photos and Android Motion Photos technologies. In this study, we introduce a straightforward methodology to evaluate and contrast the efficacy of Live/Motion Photos against traditional image-based approaches. Our findings reveal that both Live Photos and Motion Photos outperform single-frame images in common visual assisting tasks, specifically in object classification and VideoQA. We validate our results through extensive experiments on the ORBIT dataset, which consists of videos collected by visually impaired individuals. Furthermore, we conduct a series of ablation studies to delve deeper into the impact of deblurring and longer temporal crops.

## 1 Introduction

*Live Photos* and *Motion Photos*, technologies from Apple and Android, allow a single photo to function as a still image and when activated, a short video with motion and sound. These technologies leverage a background feature that continuously captures images when the Camera app is opened, regardless of whether the shutter button is pressed. When a Live/Motion Photo is taken, the device records this continuous stream of photos, capturing moments before and after the shutter press. These images are stitched into a three-second animation, complemented by optional audio recorded during the same span. Live/Motion Photos surpass video clips due to their ease of capture and standardized format. Figure 1 depicts the main three components of a Live/Motion Photo, and Figure 5 shows screenshots of the Apple iOS environment for capturing and working with Live Photos.

People with visual impairments often rely on assistive devices that provide insights about their surroundings. For instance, people with low vision often rely on magnification tools to better observe the content of interest, or those with low vision and no vision rely on on-demand technologies [BeMyEyes, 2023, BeSpecular, 2023, Khoshsirat and Kambhamettu, 2023] that deliver answers to submitted visual questions. Two fundamental computer vision tasks in these aids are object classification and video question answering (VideoQA). Object classification, though basic, is a key component of more advanced methods [Khoshsirat and Kambhamettu, 2022]. In contrast, VideoQA accurately responds to inquiries about any video, empowering visually impaired people to access information about real-world or online videos [Hosseini et al., 2022].

A significant problem with the current visual assisting technologies is the limitation of the visually impaired people to capture the desired image for these technologies. The images taken by blind people have different quality flaws, such as blurriness, brightness, darkness, obstruction, and so on [Maserat et al., 2017]. Image quality issues may make it difficult for humans and machine learning systems to recognize image content, causing the system to provide set responses, such as "unanswerable". Prior research has indicated that this can be frustrating for people with visual impairments using accessible applications, requiring extra time and effort to determine what is going wrong and get an answer [Bhattacharya et al., 2019]. Figure 2 shows a recorded video by a visually impaired user where half of the frames cover only a small portion of the object.

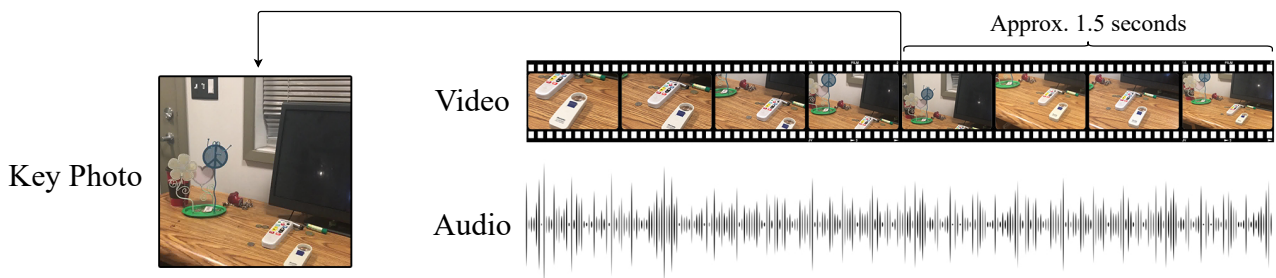

Figure 1: Apple Live Photo structure. A Live/Motion Photo consists of a key photo, a three-second-long video, and the optional corresponding audio. The key photo is the middle frame of the video, but it can be changed to another frame.

We posit that the additional contextual information provided by Live/Motion Photos can significantly enhance the ability of the assistance systems to accurately interpret and analyze the content of the images. Not only does this approach provide multiple frames for analysis, which could increase the chances of capturing a clear shot of the subject, but it also offers temporal information that can be critical for understanding dynamic scenarios. Through the course of this paper, we will present empirical evidence demonstrating how the use of Live/Motion Photos can mitigate the challenges faced by visually impaired individuals in capturing clear images.

Our contributions are as follows:

- We introduce a straightforward approach for comparing Live/Motion Photos to images.
- We evaluate state-of-the-art methods on Live/Motion Photos and images for object classification and VideoQA tasks.
- We conduct ablation studies on the impact of deblurring and varying temporal crop lengths.

## 2 Related Work

A plethora of commercial systems have been developed to empower individuals with visual impairments. These commercial systems are categorized into two distinct types: human-in-the-loop systems and end-to-end (E2E) automated systems. Human-in-the-loop systems are designed to bridge the gap between visually impaired individuals and sighted volunteers or staff members. Through these systems, users can make inquiries or seek assistance with visual tasks. Some notable examples of human-in-the-loop platforms are BeMyEyes, BeSpecular, and Aira [BeMyEyes, 2023, BeSpecular, 2023]. Contrary to human-in-the-loop systems, end-to-end systems rely on artificial intelligence and cloud computing to provide visual assistance to users. These systems do not involve human intermediaries. Examples of E2E systems include TapTapSee and Microsoft's Seeing AI.

A critical factor that determines the efficacy of these systems is the clarity and relevance of the content within the images that are sent for analysis. Given that visually impaired individuals might face challenges in capturing well-composed images, ensuring that the subject matter of the image is clear and discernible is not a trivial task. In this paper, we introduce an innovative approach to alleviate this challenge by utilizing Live Photos or Motion Photos.

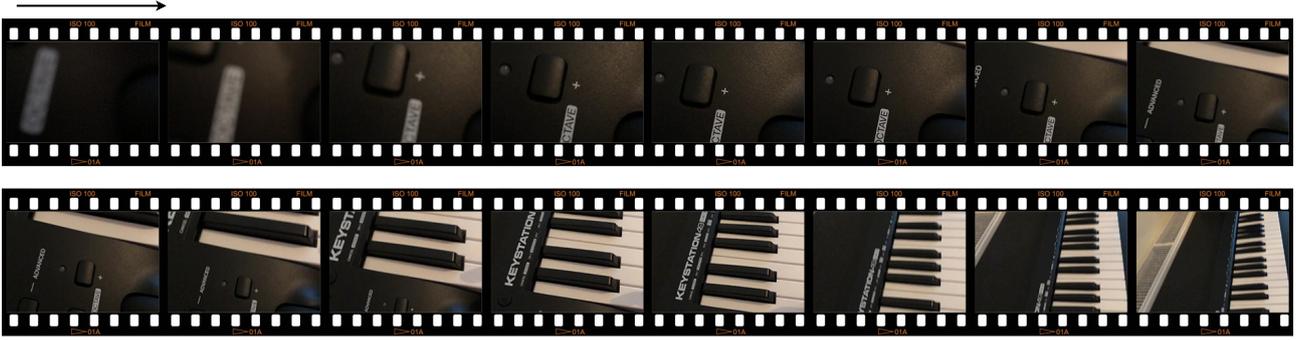

Figure 2: A visually impaired user trying to record a video of a keyboard [Massiceti et al., 2021]. Adjusting the camera field of view to cover a whole object is a challenging task for blind users. The frames are uniformly sampled, and the total video length is five seconds.

## 3 Method

Studying Live/Motion Photos poses a significant challenge due to the absence of existing datasets. The process of creating a comprehensive dataset solely from visually impaired users is laborious and complex [Massiceti et al., 2021]. To address this issue, we leverage pre-existing video datasets collected by visually impaired individuals or those containing content relevant to the daily experiences of the blind. By extracting three-second temporal crops from these videos, we simulate Live/Motion Photos for tasks such as object classification and VideoQA. This enables us to evaluate and compare the effectiveness of different methods on both simulated Live/Motion Photos and standard images.

### 3.1 Object Classification

To demonstrate the impact of Live/Motion Photos on object classification accuracy, we conduct experiments using the ORBIT dataset [Massiceti et al., 2021]. This dataset is a collection of videos recorded on cell phones by people who are blind or low-vision. The ORBIT dataset consists of 3,822 videos with 486 object categories recorded by 77 blind or low-vision people on their cell phones. Each video is meant to capture one main object, although the object may not be visible in all frames. The videos are captured in various lengths, from one second to two minutes.

To simulate Live/Motion Photos, we create short video clips with the same length as Live/Motion Photos from ORBIT and compare the performance of different image classifiers to video classifiers on these clips. To this aim, we train each image classifier on image frames of the videos and report the average classification accuracy of the frames. To evaluate the video classifiers, we train and test each method on random temporal crops of three seconds. We choose the top-performing image and video classifiers; specifically, ResNet [He et al., 2016], MViTv2 [Li et al., 2021], and EfficientNetV2 [Tan and Le, 2021] for image classification, and ViViT [Arnab et al., 2021] and MViTv2 [Li et al., 2021] for video classification. We use the same hyper-parameters and setup as in the original implementations, and the input size is fixed across all the methods. Following [Massiceti et al., 2021], we use frame accuracy as the evaluation metric for the frame-by-frame classification and video accuracy for the holistic video classification. Frame accuracy is the average number of correct predictions per frame divided by the total number of frames in a video. Video accuracy is the number of correct video-level predictions divided by the total number of videos.

Table 1 reports the object classification accuracy. The highest accuracy using images is 70.9% and achieved by EfficientNetV2-L. The results show that video classification approaches outperform frame-by-frame classification. More specifically, for Live/Motion Photos (videos of three seconds long), MViTv2 achieves an accuracy of 77.1% which is an improvement of 6.2% over EfficientNetV2-L. Since MViTv2 is designed for both image and video classification, it exhibits the benefit of using video clips over images better than other methods. Similarly, ViViT reaches an accuracy of 74.9% which is higher than EfficientNetV2-L by a margin of 4.0%. This result strongly supports the effectiveness of Live/Motion Photos over single images.

| Method | Accuracy |
|---|---|
| ResNet-152 [He et al., 2016] | 69.2 |
| MViTv2-B [Li et al., 2021] | 70.7 |
| EfficientNetV2-L [Tan and Le, 2021] | 70.9 |
| ViViT [Arnab et al., 2021] | 74.9 |
| MViTv2-B [Li et al., 2021] | 77.1 |

Table 1: Comparison of frame-by-frame to holistic object classification methods on the ORBIT test set. The top three methods use images, and the bottom two use Live/Motion Photos.

| Method | Accuracy |
|---|---|
| mPLUG [Li et al., 2022] | 28.9 |
| BEiT-3 [Wang et al., 2022] | 30.1 |
| Just Ask [Yang et al., 2021] | 34.9 |
| Singularity [Lei et al., 2022] | 38.6 |

Table 2: Results of image-based and video-based methods for the VideoQA task on the ActivityNet-QA test set. mPLUG and BEiT-3 use images, while Just Ask and Singularity use Live/Motion Photos.

## 3.2 Video Question Answering

We investigate the effectiveness of Live/Motion Photos in the VideoQA task. We compare the performance of multiple VQA methods on image frames to the performance of VideoQA methods on video clips with the same length as Live/Motion Photos. While there are numerous video question answering datasets, we choose the ActivityNet-QA dataset [Yu et al., 2019] since it contains video clips similar to the day-to-day life of people with visual impairments. The ActivityNet-QA dataset adds question-answer pairs to a subset videos of the ActivityNet dataset [Caba Heilbron et al., 2015]. The ActivityNet-QA dataset contains 5,800 videos with 58,000 human-annotated question-answer pairs divided as 3,200/1,800/800 videos for train/val/test splits. This dataset contains 200 different types of daily human activities, which is suitable for visual assisting applications.

We train image-based methods on randomly drawn frames with their corresponding question-answer pairs from the ActivityNet-QA dataset. Similarly, we train video-based methods on random temporal crops with the same length as Live/Motion Photos. We employ mPLUG [Li et al., 2022] and BEiT-3 [Wang et al., 2022] as the image-based methods and Just Ask [Yang et al., 2021] and Singularity [Lei et al., 2022] as the video-based methods for Live/Motion Photos. These methods achieve state-of-the-art accuracy in the VQA and VideoQA tasks, and their implementation code is publicly available. For each method, we re-use the original hyper-parameters that achieve the best results.

As for the evaluation criteria, we use accuracy, a commonly used criterion to measure the performance of classification tasks. For the QA pairs in the test set with size $N$, given any testing question $\mathbf{q}_i \in Q$ and its corresponding ground-truth answer $\mathbf{y}_i \in Y$, we denote the predicted answer from the model by $\mathbf{a}_i$. $\mathbf{a}_i$ and $\mathbf{y}_i$ correspond to a sentence that can be seen as a set of words. The accuracy measure is defined as:

$$Accuracy = \frac{1}{N} \sum_{i=1}^{N} \mathbf{1}[\mathbf{a}_i = \mathbf{y}_i] \qquad (1)$$

where $\mathbf{1}[\cdot]$ is an indicator function such that its output is one only if $\mathbf{a}_i$ and $\mathbf{y}_i$ are identical, and zero otherwise [Yu et al., 2019]. We follow previous evaluation protocols for open-ended settings [Yang et al., 2021, Yu et al., 2019, Lei et al., 2022] and use a fixed vocabulary of training answers.

Table 2 reveals the results of our experiments for the VideoQA task. The highest accuracy for image-based approaches is 30.1% and achieved by BEiT-3. Both VideoQA methods outperform the VQA methods. More specifically, using Live/Motion Photos, Singularity achieves the highest accuracy of 38.6%, which is more than 8% higher than BEiT-3 accuracy. Similarly, Just Ask reaches an accuracy of 34.9% which is 4.8% higher than BEiT-3.

The outcomes of our experiments in object classification and VideoQA confirm the benefit of using Live/Motion Photos over images.

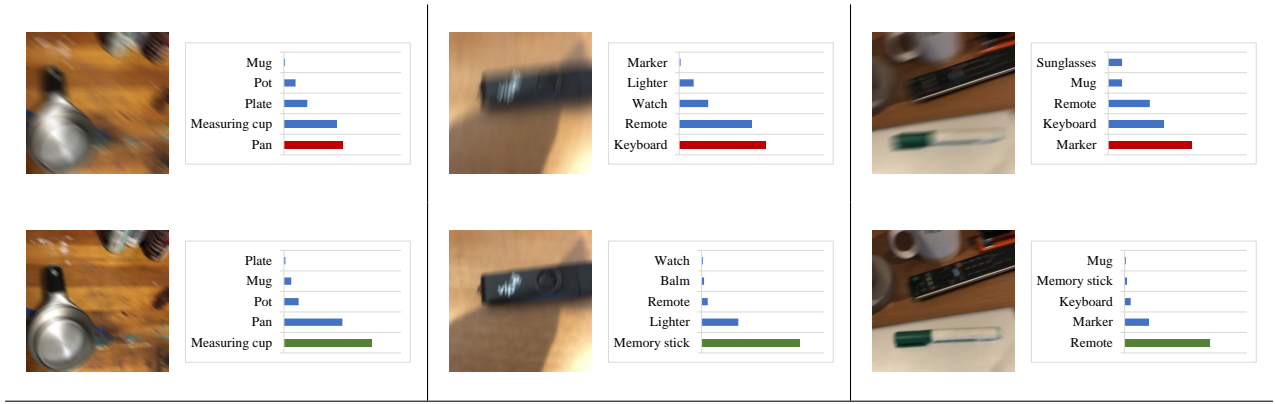

Figure 3: Sample video frames from ORBIT dataset with their corresponding model output. **Top:** Original frame. **Bottom:** After deblurring. Deblurring enhances the precision of model predictions.

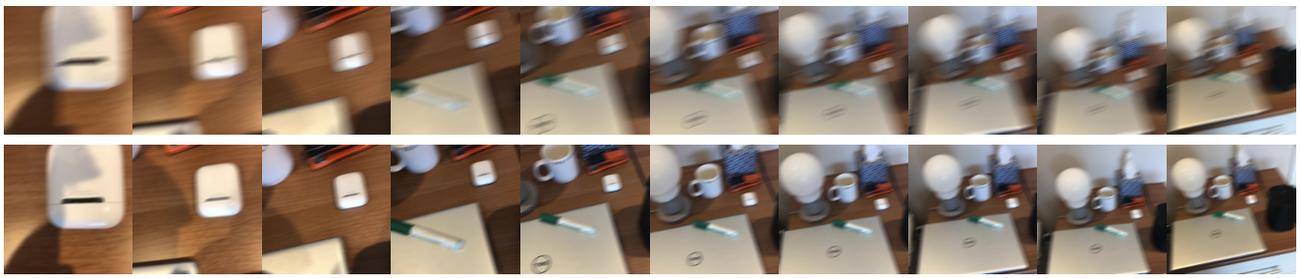

Figure 4: Ten uniformly sampled frames from a random video in ORBIT, before and after deblurring. **Top:** Original video. **Bottom:** After deblurring. Deblurring tends to provide greater benefits to frames containing smaller objects.

## 4 Deblurring Impact

Blurring is a prevalent issue in images and videos captured by individuals with visual impairments [Chiu et al., 2020], and this issue can adversely affect the efficacy of assistive technologies. In this section, we undertake a systematic investigation to discern the potential benefits of deblurring on the accuracy of object classification and VideoQA. For the deblurring process, we employ the FGST method [Lin et al., 2022], a state-of-the-art video deblurring algorithm that amalgamates Transformer modules with optical flow estimation. We then proceed to apply FGST on two datasets, namely ORBIT and ActivityNet-QA, to deblur the visual content. With the deblurred datasets, we replicate the experiments as outlined in Section 3.1 and Section 3.2.

The outcomes of this investigation are tabulated in Table 3. The table segregates the results into two categories - the upper portion presents the outcomes for object classification, while the lower portion provides the results for VideoQA. The empirical findings demonstrate that the maximum enhancement in accuracy is 2.6%, which is attained by the Just Ask method, whereas the minimum improvement is documented at 1.7% by the MViTv2-B method. Furthermore, for a more illustrative understanding, Figure 3 showcases a selection of frames along with the corresponding model outputs prior to and subsequent to the deblurring process. This visualization facilitates a comparison of the quality and detail in the frames. Additionally, Figure 4 presents a compilation of frames extracted from a deblurred video, providing a visual representation of the enhancements achieved through the deblurring process.

| Method | Without Deblurring | With Deblurring |
|---|---|---|
| ViViT [Arnab et al., 2021] | 74.9 | 76.9 (+2.0) |
| MViTv2-B [Li et al., 2021] | 77.1 | 78.8 (+1.7) |
| Just Ask [Yang et al., 2021] | 34.9 | 37.5 (+2.6) |
| Singularity [Lei et al., 2022] | 38.6 | 40.9 (+2.3) |

Table 3: The impact of deblurring Live/Motion Photos. **Top:** Object classification on the ORBIT test set. **Bottom:** VideoQA on the ActivityNet-QA test set.

| | Accuracy | | | | |
|---|---|---|---|---|---|
| Method | Live/Motion Photo =3s | Short <15s | Medium 15s> and <30s | Long >30s | All Frames |
| ViViT [Arnab et al., 2021] | 74.9 | 75.8 | 76.6 | 77.1 | 76.7 |
| MViTv2-B [Li et al., 2021] | 77.1 | 77.9 | 78.4 | 79.0 | 78.5 |
| Just Ask [Yang et al., 2021] | 34.9 | 36.0 | 36.9 | 37.8 | 37.0 |
| Singularity [Lei et al., 2022] | 38.6 | 39.6 | 40.4 | 41.1 | 40.6 |

Table 4: The results of top-performing methods with different temporal crop lengths. **Top:** Object classification on the ORBIT test set. **Bottom:** VideoQA on the ActivityNet-QA test set. Videos shorter than a targeted crop size are not included in that group.

## 5 Temporal Length Impact

Although Live/Motion Photos are limited to three seconds, it is possible for other applications to implement the same technology but without the capturing limitations. Therefore, in this section, we study the effect of video length on accuracy for object classification and VideoQA tasks. To this aim, we evaluate the video-based methods on three temporal crop size ranges. The 'Short' crop range is the random crops of shorter than 15 seconds, the 'Medium' range is between 15 to 30, and the 'Long' range is longer than 30 seconds. Videos that are shorter than a targeted crop size are not included in that group. Additionally, we evaluate the methods on the whole dataset using all the available frames in the videos. We do not use videos shorter than the required minimum length for the Medium and Long ranges. We use the same setup in Sections 3.1 and 3.2.

For object classification, we employ ViViT [Arnab et al., 2021] and MViTv2-B [Li et al., 2021] and evaluate them on the ORBIT dataset [Massiceti et al., 2021]. The top two methods in Table 4 report the results for object classification using different video lengths. For MViTv2-B, the lowest accuracy is 77.1%, achieved by using Live/Motion Photos, and the highest accuracy is 79.0%, achieved using the longest video crops. For both methods, adding more frames helps improve the accuracy. The accuracy of using all the frames gets slightly worse due to the addition of shorter videos. Since having more frames reveals more data about an object, the longer crops reach higher accuracies.

For VideoQA, we employ Just Ask [Yang et al., 2021] and Singularity [Lei et al., 2022] and train and test them on the ActivityNet-QA dataset [Yu et al., 2019]. The bottom two methods in Table 4 report the results for different video lengths in VideoQA. The lowest accuracy for Singularity is 38.6% by using Live/Motion Photos, and 41.1% is the highest accuracy by using all the frames.

The findings from our ablation study reveal that while there is a positive correlation between the length of video clips and the enhancement in accuracy, the incremental accuracy attained through longer video clips, as compared to Live/Motion Photos, is not significant in contrast to single images. This implies that Live/Motion Photos, constrained to a duration of three seconds, are capable of furnishing a substantial improvement in accuracy that is deemed sufficient for a majority of applications.

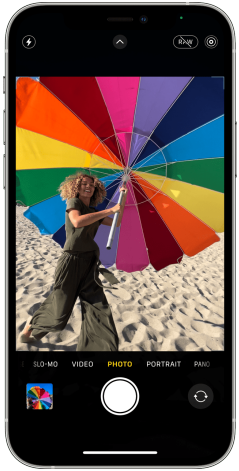 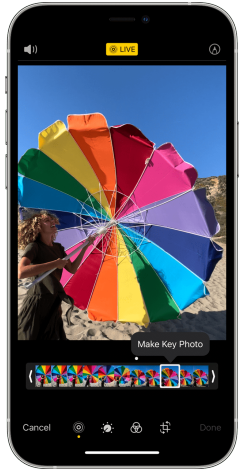 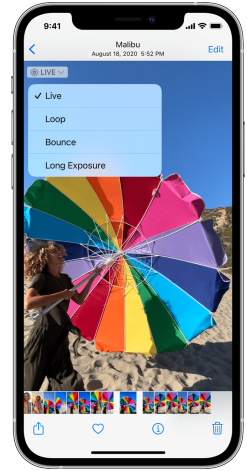

(a) Capturing a Live Photo  (b) Changing the key photo  (c) Editing a Live Photo

Figure 5: Three screenshots showcasing Live Photos functionality on Apple iOS [Apple, 2021].

## 6 Conclusion and Future Directions

Despite significant recent developments, visual assistance applications are still in need of improvement. Current machine learning methods designed to help visually impaired people suffer from the low quality of the images taken by the end users.

In this paper, we made multiple contributions to improving existing methods for visual assisting. We introduced a simple way to evaluate the performance of Live/Motion Photos compared to single images. We employed this approach to show that Live/Motion Photos achieve higher accuracy in common visual assisting tasks. Our experiment revealed that Live/Motion Photos perform better than images in object classification and VideoQA tasks. In addition, we further studied the effect of longer temporal crops and showed how deblurring can improve accuracy.

In future research, it is essential to carry out user studies involving visually impaired individuals. This information will guide us in refining our method, ensuring it is not only technically robust but also practically beneficial for the intended users.